\documentclass[runningheads]{llncs}

\pdfoutput=1

\usepackage{eccv}
\usepackage{eccvabbrv}

\usepackage{graphicx}
\usepackage{booktabs}
\usepackage{multirow}
\usepackage{orcidlink}
\usepackage[accsupp]{axessibility}

\usepackage{hyperref}

\newcommand{\ours}{MDiFF\xspace}

\begin{document}

\title{MDiFF: Exploiting Multimodal Score-based Diffusion Models for New Fashion Product Performance Forecasting}

\titlerunning{MDiFF: Multimodal Score-based Diffusion Models for NFPPF}

\author{Andrea Avogaro\inst{1}\orcidlink{0000-0003-2244-2785} \and
Luigi Capogrosso\inst{1}\orcidlink{0000-0002-4941-2255} \and
Franco Fummi\inst{1}\orcidlink{0000-0002-4404-5791} \and \\
Marco Cristani\inst{1}\orcidlink{0000-0002-0523-6042}
}

\authorrunning{A.~Avogaro et al.}

\institute{Dept. of Engineering for Innovation Medicine, University of Verona, Verona, Italy
\email{name.surname@univr.it}}

\maketitle

\begin{abstract}
The fast fashion industry suffers from significant environmental impacts due to overproduction and unsold inventory.
Accurately predicting sales volumes for unreleased products could significantly improve efficiency and resource utilization.
However, predicting performance for entirely new items is challenging due to the lack of historical data and rapidly changing trends, and existing deterministic models often struggle with domain shifts when encountering items outside the training data distribution.
The recently proposed diffusion models address this issue using a continuous-time diffusion process.
This allows us to simulate how new items are adopted, reducing the impact of domain shift challenges faced by deterministic models.
As a result, in this paper, we propose \ours{}: a novel two-step multimodal diffusion models-based pipeline for New Fashion Product Performance Forecasting (NFPPF).
First, we use a score-based diffusion model to predict multiple future sales for different clothes over time.
Then, we refine these multiple predictions with a lightweight Multi-layer Perceptron (MLP) to get the final forecast.
\ours{} leverages the strengths of both architectures, resulting in the most accurate and efficient forecasting system for the fast-fashion industry at the state-of-the-art.
The code can be found at \url{https://github.com/intelligolabs/MDiFF}.


\end{abstract}

\section{Introduction} \label{sec:sec_intro}

The fast fashion industry represents the second most pollutive industry in the world, responsible for 79 trillion liters of water consumed and 92 million tonnes of waste produced per year~\cite{niinimaki2020environmental}, contributing 8\% of all carbon emissions and 20\% of all global wastewater~\cite{bailey2022environmental}.
Being able to predict more precisely the sales volumes for an unreleased product could represent a significant step toward making this market more efficient, reducing the use of resources for production and especially minimizing leftover unsold inventory.
While forecasting time series with a known historical past has been analyzed extensively~\cite{ahmed2010empirical,lim2021time}, very little attention has been paid to a much more practical and challenging scenario: forecasting new products that the market has not seen before.

Specifically, this problem, known as \textit{New Fashion Product Performance Forecasting (NFPPF)}~\cite{skenderi2024well}, is far from trivial since unreleased products have no past sales data available.
Thus, it is necessary to retrieve valuable information from the available data, which may be technical specifications of the product (such as color, type, material), release period, or interest shown for similar products in the past~\cite{skenderi2024well}.

\paragraph*{\textbf{Motivation for this paper.}}
Due to the rapidly changing nature of fashion trends, determining what is considered fashionable or outdated can be challenging.
This makes it difficult to accurately predict the market performance of a specific item and identify the key factors that influence its popularity.
Traditional deterministic forecasting models have demonstrated reasonable performance in specific situations. 
However, they are limited by their assumption that the characteristics of past-season products are directly applicable to new items, which often exhibit distinct features.
This leads to inaccurate predictions due to the shift in the characteristics of the input data.

\begin{figure*}[t!]
    \centering
    \includegraphics[width=\linewidth]{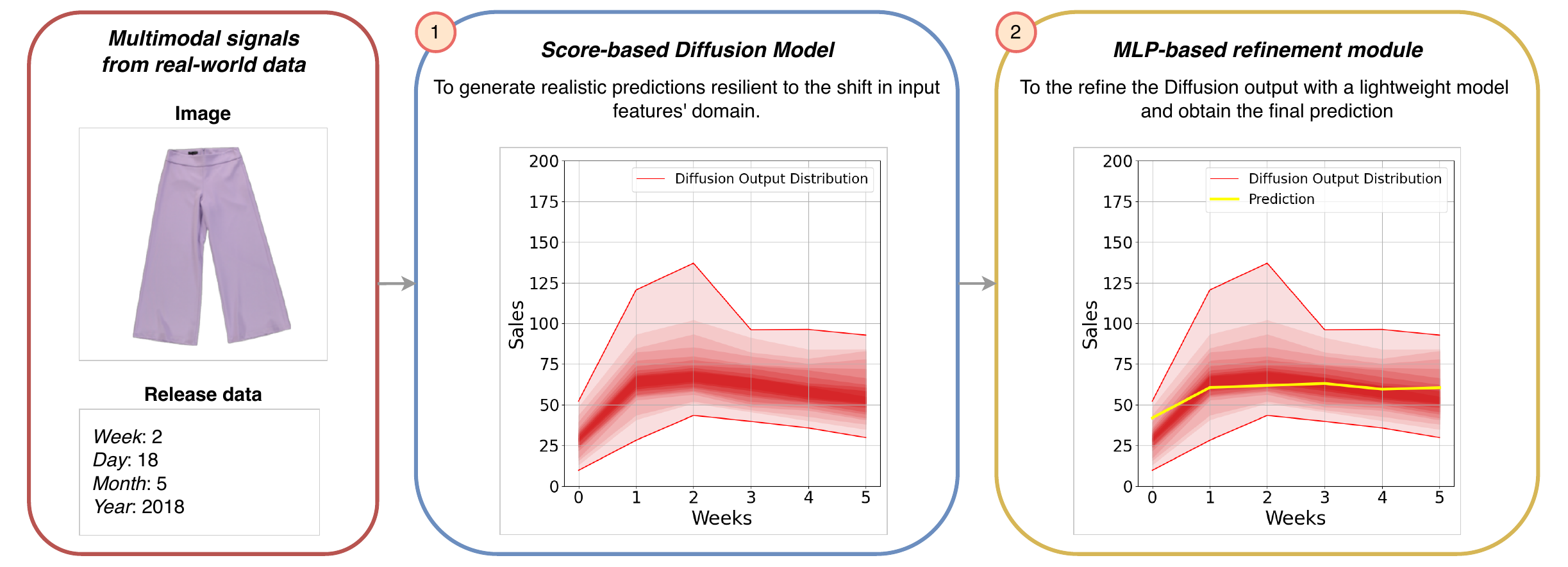}
    \caption{
    \ours{}: a two-stage pipeline for NFPPF.
    Starting from multiple signals of a single fashion product, we build a multimodal score-based diffusion model to generate an initial prediction of the sales, addressing potential objects with features beyond the training distribution.
    Then, we refine the Diffusion output using a lightweight MLP in order to obtain the final prediction.
    }
    \label{fig:fig_figure_1}
\end{figure*}
Recently, Denoising Diffusion Probabilistic Models (DDPMs)~\cite{ho2020denoising}, or diffusion models in general, have impressed with their realistic image and video generation capabilities.
Moreover, DDPMs have demonstrated promising results even in the context of time series analysis~\cite{lin2023diffusion}.
Specifically, DDPMs implicitly learn the probability distributions of data, such as images~\cite{capogrosso2024diffusion,girella2024leveraging}, or fashion sales, as we demonstrate in this paper.
As a result, they are unaffected by the feature's domain shift issue due to the nature of the diffusion process~\cite{sohl2015deep}.
Indeed, a DDPM is trained gradually to denoise Gaussian noise; therefore, it does not use explicit features extracted from a specific sample as input.
This characteristic is critically important, making DDPMs an ideal tool for NFPPF.

In particular, when a DDPM comes across a sample with features that aren't in its training data (a common occurrence in the fast-fashion industry), its denoising diffusion process naturally keeps its predictions within the real sales distribution.
Furthermore, even when a DDPM is conditioned on extracted features, the model's predictions don't diverge significantly~\cite{yang2023diffusion}.
On the contrary, deterministic approaches don't have this kind of reliability because they directly link input features to predicted sales, which can result in wrong predictions for new feature combinations.

\paragraph*{\textbf{Architectectural innovations.}}
In this paper, we introduce \ours{}, a two-stage architecture shown in Figure~\ref{fig:fig_figure_1} specifically crafted to tackle NFPPF.
Starting from VISUELLE~\cite{skenderi2022multi,skenderi2024well}, we first train a multimodal score-based diffusion model that learns how to generate samples from the true sales distribution.
A second refining model, based on a Multi-layer Perceptron (MLP), is then used to refine the DDPM prediction.
DDPMs, by nature, generate output from Gaussian noise, guaranteeing a fundamental property, which is the generation of non-deterministic samples.
In order to be able to control this behaviour better and have more stable results, what we do is generate multiple sales signals for the same object. Specifically, 50 samples are generated and given as input to the refinement model, which subsequently generates the final sales prediction.
This strategy ensures that the MLP receives in-distribution data that more closely matches the actual sales data distribution, leading to improved pipeline reliability.
We then evaluated \ours{} on VISUELLE, a dataset specifically built to serve as a benchmark specifically designed for NFPPF.
To summarize, the main contributions of our work are as follows:
\begin{itemize}
\item We propose \ours{}, a novel two-stage pipeline for NFPPF that utilizes diffusion models to generate initial predictions, even for cases with features beyond the training distribution.
\item These initial predictions are then refined through a lightweight MLP model to produce our final forecast.
\item We conducted experiments on the VISUELLE benchmark, demonstrating that \ours{} surpasses existing state-of-the-art methods in NFPPF.
\end{itemize}

The paper is organized as follows.
Related works are reported in Section~\ref{sec:sec_related}.
Our proposed method is described in Section~\ref{sec:sec_method}, followed by experimental results in Section~\ref{sec:sec_experiments}.
Finally, conclusions are drawn in Section~\ref{sec:sec_conclusions}.

\section{Related Works} \label{sec:sec_related}

This section provides an overview of existing NFPPF approaches and discusses different available datasets for this task.

\subsection{New Fashion Product Performance Forecasting (NFPPF) Problem} \label{subsec:subsec_nfppf_problem}
Research on NFPPF utilizing machine and deep learning techniques is still in its early stages but is growing rapidly.
The problem was first introduced in~\cite{ren2017comparative}, which analyses the importance of predicting the demand for fast fashion supplies in order to improve efficiency in the production chain.
This study explores different classical statistical models to tackle this problem and gives some preliminary results applied to real data.

Subsequently,~\cite{craparotta2019siamese} aimed to mimic expert judgment by training a Convolutional Neural Network (ConvNet) to identify and extract visual product characteristics and then using classical Machine Learning (ML) approaches to produce the sales.
Specifically, the extracted features from images are then fed into a k-Nearest Neighbors (k-NN) algorithm model to produce the final predictions.

Similarly,~\cite{singh2019fashion} tried to explore different algorithms like gradient boosting, random forest, and k-NN to set a defined machine-learning baseline for the problem.

Then,~\cite{singh2019fashion} showed how Deep Neural Networks (DNNs) are outperforming this baseline, proposing two solutions based on Feed-Forward Networks and Long Short-Term Memory (LSTM).
As input, the authors formed a multimodal signal derived from fixed product characteristics like category, color, and fabric, as well as dynamic factors such as discounts and promotions, without any visual information about the garment in order to reduce the complexity of the model.

Similarly, in~\cite{ekambaram2020attention}, the authors use an architecture based on Recurrent Neural Networks (RNNs), including more signals like past sales, images, textual embeddings, and discounts.
The model also employed a soft-attention mechanism to determine the most influential factors in predicting sales.
However, the autoregressive nature of the model resulted in identical predictions for products in different seasons.
Unfortunately, the code and data behind this research remain proprietary and inaccessible to the public.

However, the main problem is the lack of a common dataset on which comparative tests can be conducted to investigate which models may be of real practical relevance for adopting these systems.
Publicly available datasets for fashion forecasting like~\cite{hmdataset} consider diverse applications dissimilar from NFPPF.
Typically, these datasets are constructed and used to predict general fashion trends, determined by the popularity of various brands' products on social media platforms such as Instagram.
In addition, no actual sales figures are considered, only trends.

For this reason, in~\cite{skenderi2022multi,skenderi2024well}, the authors released VISUELLE, a novel dataset containing exogenous signals like images, attributes, and Google trends related to items sold by a fast-fashion company in over 100 physical shops with true sales signals.

Based on this benchmark, in~\cite{skenderi2024well}, the authors propose GTM, an encoder-decoder Transformer-based architecture incorporating all the multimodal data offered by the dataset as input of the model.
The encoder is fed with Google Trend signals, while the decoder receives a set of features extracted from images, textual descriptions, and temporal information of the article (release date).
This approach effectively extends previous work by exploiting a more powerful architecture to extract insights from exogenous signals, demonstrating the effectiveness of Google Trends information concerning NFPPF.

Finally,~\cite{joppi2022pop} proposes a novel data-centric pipeline that generates synthetic historical data for every item based on images of the same category downloaded by Google Images, then fed into GTM, showing promising results.
Unfortunately, we did not include these results in our paper because the code and the images used by the authors are unavailable.

\section{Methodology} \label{sec:sec_method}

In this section, we introduce the problem formalization (Section~\ref{subsec:subsec_problem_formalization}), our score-based diffusion model (Section~\ref{subsec:subsec_score_dm}), the methodology used to guide sample generation (Section~\ref{subsec:subsec_multimodal_conditioning}), and the MLP-based refinement stage (Section~\ref{subsec:subsec_refinement_module}).

\subsection{Problem Formalization} \label{subsec:subsec_problem_formalization}
Given a new product $j$, we want to predict $y\in\mathbb{R}^{W}$ expressed as the performance vector in terms of sales in an interval of $W$ weeks since its release date.

For every $j$, a set of 2 attributes is given: an image of the product $i_j \in \mathbb{R}^{w\times h\times 3}$ with $w = h =256$ and release date $t_j\in\mathbb{R}^4$ composed by four digits representing day, week, month and year of release.

\subsection{Our Score-Based Diffusion Model} \label{subsec:subsec_score_dm}
Score-based diffusion models~\cite{song2020score} generalize DDPMs~\cite{ho2020denoising} generative models trained to reverse a discrete-time diffusion process.
A Gaussian noise diffusion process, also known as the \textit{forward process}, can be summarized as a chain of steps in which Gaussian noise is progressively added to the initial distribution, as described by the following equations:
\begin{equation}
    q(x^{1},...,x^{T}|x^{0}=y)=\prod_{t=1}^{T}q(x^{t}|x^{t-1})\;,
\end{equation}
\begin{equation}
    q(x^{t}|x^{t-1}):=\mathcal{N}(\sqrt{1-\beta_{t}}x_{t-1},\beta_{t}I)\;,
\end{equation} where $q(x^T)\approx \mathcal{N}(0,1)$, $y=q(x^0)$ is the true data distribution, $\beta_t$ is the variance of the additive noise, and $t\in [0,T]$ represents the number of noising steps defined a prior. 

A model $p_\theta$ is then trained to reverse the diffusion process by gradually removing noise, also known as the \textit{backward process}, to restore the initial distribution.
Specifically, the backward process is formalized as follows:
\begin{equation}
    \begin{split}
    p_\theta(x^{t-1}|x^t,c) = \mathcal{N}(x^{t-1}; \mu_\theta(x^t,t) &\\  + s\sigma_t^2\nabla_{x^t}\text{log}p(x^t|x^0), \sigma_t^2 I)\;,
    \end{split}
\end{equation}
where $\sigma$ is the variance for each timestep, and $s$ is the parameter that controls the strength of the conditioning.

\begin{figure*}[t!]
    \centering
    \includegraphics[width=\linewidth]{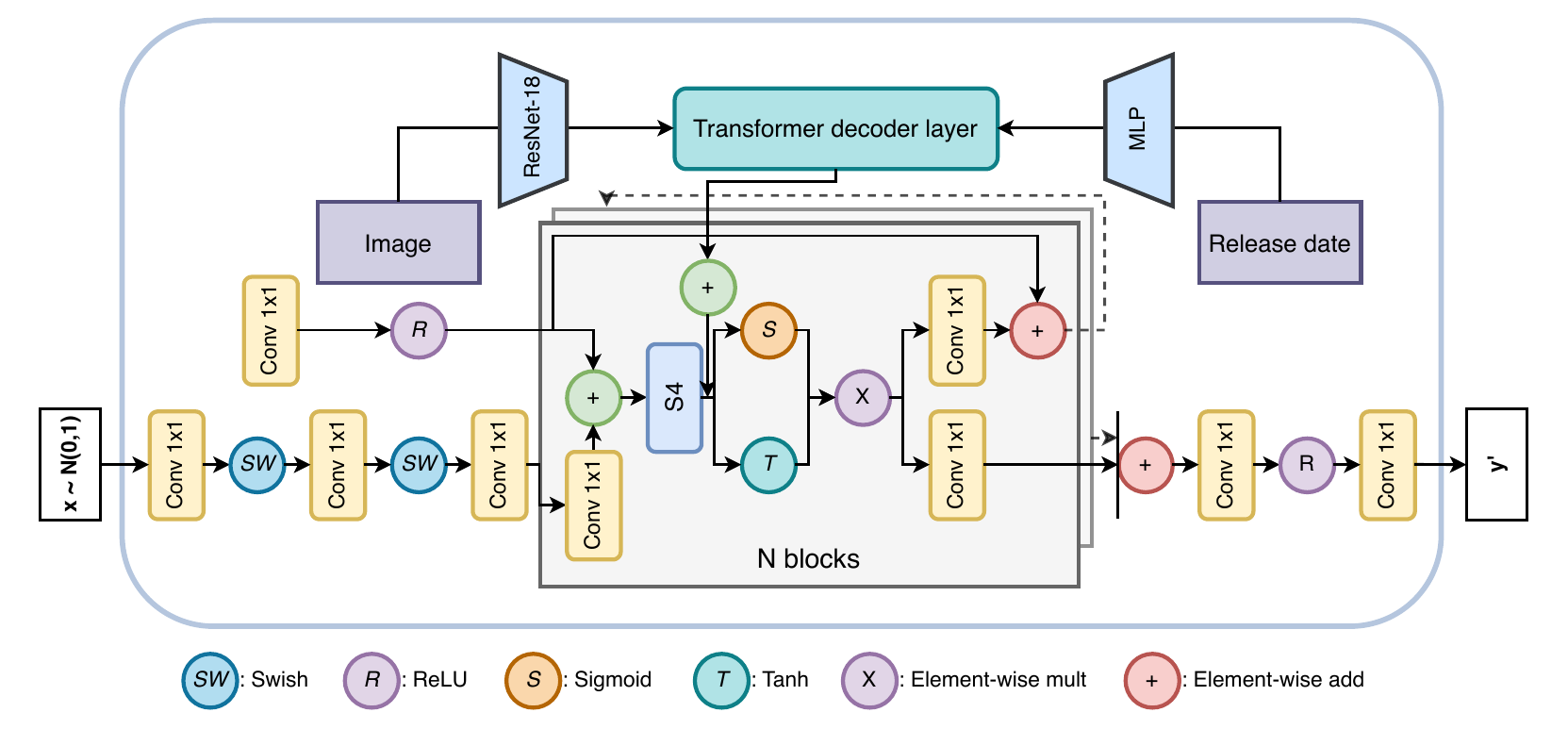}
    \caption{
    An overview of our multimodal score-based diffusion model.
    The diffusion basic block is taken from TS-Diff~\cite{kollovieh2024predict} (grey square), modified to be injected with the output of the transformer decoder layer, a module responsible for producing an embedding representing the two modalities of input related to the item.
    Each block contains two outputs: one for the subsequent block and another for a skip connection.
    The summation of all skip connections forms the model's final output.
    The primary component of each block is typically an S4 block~\cite{gu2021efficiently}, chosen by the authors of~\cite{kollovieh2024predict} for its efficiency when it comes to time series and structured data.
    The input of the MDiFF is noisy data, and the output is the denoised sample.}
    \label{fig:figure_2}
\end{figure*}
Specifically, Figure~\ref{fig:figure_2} shows the architecture of our multimodal score-based diffusion model.
The network is a stack of multiple $N$ blocks.
Every block has two outputs, one for the next block and one for skip connection.
The summation of all the skip connections represents the actual output of the model.
Every block is mainly composed of an S4 block~\cite{gu2021efficiently}.
The \textit{multimodal conditioning} $c_j$ is directly added to the output of the S4 block, and the cross-attention is implemented using a Transformer decoder layer~\cite{vaswani2017attention}.

\subsection{Multimodal Conditioning} \label{subsec:subsec_multimodal_conditioning}
To guide the generation of the future sales of the $j$-th product by the Diffusion Model, we use multimodal data composed of images $i_j$ of the product and release date $t_j$, as described in Section~\ref{subsec:subsec_problem_formalization}.

We train two different encoders $I_{\theta}, T_{\theta}$ to extract features from the images and release dates respectively.
Then, these features are used to produce the conditional embedding through a cross-attention mechanism defined as:
\begin{align}
    c_j &= \text{Softmax}\left(\frac{Q_j K_j^T}{\sqrt{d_k}}\right) V_j\;,
\end{align}
where $K_j=V_j=I_{\theta}(i_j)$, $Q_j=T_{\theta}(t_j)$ and $\sqrt{d_k}$ is the dimensionality of the number of features of the embeddings.

We designed the conditioning module of \ours{} to weight, through the attention mechanism, the image embedding $I_{\theta}(i_j)$ with the features $T_{\theta}(t_j)$ obtained from the temporal information given by the release date.
The idea behind this architectural choice came from the fact that every visual feature of the item has to be considered with respect to the fashionable concept of the current season to guide the reverse diffusion process effectively.
To effectively and efficiently use the cross-attention mechanism, we used a Transformer decoder layer to serve this scope.
More details about the implementation of the various encoders are reported in Section~\ref{subsec:subsec_exp_setup}

\subsection{MLP-Based Diffusion Outputs Refinement} \label{subsec:subsec_refinement_module}
We approach the refinement stage as a regression task to predict a continuous output value.
The model is designed to reduce the dimensionality of all 50 predictions of the Diffusion Model, analyzing both the feature and temporal dimensions. 
Given $x \in \mathbb{R}^{WxN}$ the set of predictions of the Diffusion Model, $y$ the true sales signal and $\hat{y}$ the output of the model with $y,\hat{y} \in \mathbb{R}^{1xW}$, the model is defined as follows:
\begin{equation}
    \begin{split}
    x_t = \phi_t(x)\;,&\\
    \hat{y} = \phi_n(x_t')\;,
    \end{split}
\end{equation}
where $N=50$ is the number of generated samples of the Diffusion Model, $W=6$ is the number of weeks of prediction, $\phi_t(x)$ the temporal MLP defined as $\phi_t(x) = W_t x + B_t$ and $\phi_n(x) = W_n x + B_n$ the MLP that regress the final sales signal.

Specifically, for $\phi_t(x)$, $W_t \in \mathbb{R}^{WxW}$ and $B_t \in \mathbb{R}^{WxN}$; on the other hand, for $\phi_n(x)$ we have $W_n \in \mathbb{R}^{1xN}$ and $B_n \in \mathbb{R}^{1xW}$.
In this case, the dimensionality $W$ is the number of weeks to predict, while $W_t,W_n$ are the weight matrices of the two MLPs.
Specifically, we utilized an MLP network trained with a Mean Squared Error (MSE) loss function denoted as $\mathcal{L}{\text{MSE}}$.
The mean squared error loss measures the dissimilarity between the predicted output and the ground truth.
Mathematically, it is defined as:
\begin{equation}
    \mathcal{L}_{\text{MSE}}(y, \hat{y}) = \frac{1}{N} \sum_{i=1}^{N} (y_i - \hat{y}_i)^2\;,
\end{equation}
In detail, $y_i$ denotes the true output value for sample $i$, while $\hat{y}_i$ signifies the predicted output value for sample $i$.
More details about the structure of the refinement module are explained in Section~\ref{subsec:subsec_exp_setup}.
Instead, in Section~\ref{subsec:subsec_qualitative}, we motivate the rationale for using a refinement module.

\section{Experiments} \label{sec:sec_experiments}

Here, we first present the experimental setup (Section~\ref{subsec:subsec_exp_setup}). 
Then, we show our quantitative results among eight state-of-the-art competitors (in Section~\ref{subsec:subsec_quantitative}), and we argue regarding the necessity of a refinement module that we designed as an MLP (in Section~\ref{subsec:subsec_qualitative}).
Finally, we report the ablation study in Section~\ref{subsec:subsec_ablation} to validate our proposal further.

\subsection{Experimental Setup} \label{subsec:subsec_exp_setup}

\paragraph*{\textbf{Implementation details.}}
The backbone of our model is a TS-Diff~\cite{kollovieh2024predict} architecture, a multi-purpose score-based diffusion model developed for predicting, reconstructing, and refining \textit{univariate} time series data.
TS-Diff consists of a series of $M=4$ S4 Blocks~\cite{gu2021efficiently} layers, connected via skip connections.
We have extended the model to be conditioned as described in Section~\ref{subsec:subsec_multimodal_conditioning}, using a Transformer decoder layer~\cite{vaswani2017attention} to implement the multi-head cross-attention used to ensembling the two embeddings.

\noindent{\textbf{Image encoder.}}
We used as Image Encoder $I_{\theta}$ a ResNet-18~\cite{he2016deep} pre-trained on ImageNet-1K~\cite{deng2009imagenet}.
We substituted the last two layers of the model with a Conv1D and a Linear to reduce the dimensionality of the features extracted, obtaining a tensor $I_{\theta}(i_j) \in \mathbb{R}^{C \times W}$, with $C$ channels equal to 64 and $W$ forecasting horizon of six weeks.

\noindent{\textbf{Temporal encoder.}}
The Temporal Encoder $T_{\theta}$ comprises four different MLPs that expand the dimension from 1 to $C$.
The output of this model is then concatenated along the channel dimension and fed into another MLP that reduces the feature number from $4C$ to $C$, resulting in $T_{\theta}(t_j) \in \mathbb{R}^{C}$.

\noindent{\textbf{MLP-based refinement.}}
The refinement network is based on two MLPs working on different dimensions of the input tensor.
The first part is a stack of five Linear layers working on the temporal dimension of the input (\emph{i.e.}, the six weeks of prediction), expanding and compressing the feature space to match the same dimensionality of the input.
The second part comprises three other linear layers, operating on the sample's dimensionality.
These layers gradually reduce and compress the $N=50$ predictions to achieve the actual forecasting.
We used the $ReLU$ as an activation function, with a skip connection between the input tensor and the output of the first MLP.

\paragraph*{\textbf{Dataset description.}}
We used the VISUELLE fast-fashion dataset~\cite{skenderi2022multi,skenderi2024well} to test our proposal.
The dataset provides a comprehensive collection of fashion products and consumer behavior data.
It encompasses three primary components: product information, customer data, and market trends. 

Product information includes detailed descriptions of individual items.
This involves visual representations in the form of high-resolution images showcasing the product on a plain background.
Additionally, textual attributes such as product category, color, fabric, and release date are provided.

Customer data offers meaningful insights into consumer preferences and purchasing habits.
It contains anonymized information about a large number of customers, including their purchase history, specific items purchased, purchase dates, and the stores where purchases were made. 

Finally, market trend data is incorporated into the Google Trends time series.
This information tracks the popularity of product attributes like color, category, and fabric over time, providing valuable insights into consumer interest and demand fluctuations.

Specifically, the VISUELLE dataset provides purchase information of 667K users, containing data regarding 5,577 products exposed in 100 shops of Nunalie, an Italian fast-fashion company. 
The dataset contains 5,080 samples for the training set and 497 for the testing set.
The dataset does not serve a proper validation set to evaluate during the training procedure model.
Therefore, our model was evaluated during the training on the test set.

Since we aim to put ourselves in a more challenging scenario, we rely only on the image and the release date as multimodal conditioning for each product, skipping the Google Trends and description ground truth available.
Specifically, the fact that \ours{} is conditioned only with image and release date is definitely a pro since these data types are extremely easy to find, minimizing the need for annotations by object operators and much more applicable on a large scale automatically in the fast fashion market.

Paired with every item, the dataset gives the sum of the sales across all 100 shops of the specific product in the 12 weeks after the release date.
Following the evaluation protocol of~\cite{joppi2022pop,skenderi2024well}, we only predict the first 6 values of the available interval.

\paragraph*{\textbf{Evaluation metrics.}}
The Mean Average Error (MAE) and Weighted Absolute Percentage Error (WAPE)~\cite{hyndman2008forecasting}, \emph{i.e.}, the two main metrics representing the quality of the forecasting, are used to evaluate \ours{}.
Formally, they are defined as:
\begin{equation} \label{eq:eq_mae}
    \text{MAE}=\frac{\sum_{t=0}^T |y_t-\hat{y_t}|}{T}\;,
\end{equation}
\begin{equation} \label{eq:eq_wape}
    \text{WAPE}=\frac{\sum_{t=0}^T |y_t-\hat{y_t}|}{\sum_{t=0}^T y_t}\;,
\end{equation}
where $y$ represents the actual values of the time series, $\hat{y}$ represents the forecasted values, and $T$ represents the total number of observations in the time series.

\paragraph*{\textbf{Training details.}}
All the code is implemented in PyTorch~\cite{paszke2019pytorch}.
For the multimodal score-based diffusion model, we train the network for 500 epochs, with a learning rate of $1\times{}10^{-3}$, a weight-decay of $5\times{}10^{-4}$, using AdamW~\cite{loshchilov2017decoupled} as an optimizer, on a NVIDIA RTX 4090.
On the other hand, for the MLP networks, a Bayesian algorithm was used to search for the best training hyperparameters of the refinement MLP network.

\subsection{Quantitative Results} \label{subsec:subsec_quantitative}
\begin{table*}[t!]
    \caption{Quantitative results of \ours{} expressed in terms of WAPE and MAE, described in Equation~\ref{eq:eq_wape} and Equation~\ref{eq:eq_mae}, respectively.
    In \textbf{bold}, the best results.
    \underline{Underlined}, the second best.}
    \begin{tabular}{l|cccc|cc}\toprule
    \multirow{2}{*}{\textbf{Model}} &
    \multirow{2}{*}{\textbf{Image}} &
    \multirow{2}{*}{\begin{tabular}{@{}c@{}}\textbf{Temporal} \\ \textbf{Condition}\end{tabular}} &
    \multirow{2}{*}{\textbf{Description}} &
    \multirow{2}{*}{\begin{tabular}{@{}c@{}}\textbf{Google} \\ \textbf{Trends}\end{tabular}} &
    \multirow{2}{*}{\textbf{WAPE $\downarrow$}} &
    \multirow{2}{*}{\textbf{MAE $\downarrow$}} \\
    & & & & & & \\
    \midrule
    Attribute k-NN~\cite{ekambaram2020attention} & & &\checkmark & &59.8 &32.7 \\
    Image k-NN~\cite{ekambaram2020attention} &\checkmark & & & &62.2 &34 \\
    Attr + Image k-NN~\cite{ekambaram2020attention} &\checkmark & &\checkmark & &61.3 &33.5 \\
    GBoosting~\cite{friedman2001greedy} &\checkmark &\checkmark & & &64.1 &35 \\
    GBoosting+G~\cite{friedman2001greedy} &\checkmark &\checkmark & &\checkmark &63.5 &34.7 \\
    Cat-MM-RNN~\cite{ekambaram2020attention} & &\checkmark &\checkmark &\checkmark &63.3 &34 \\
    X-Att-RNN~\cite{ekambaram2020attention} &\checkmark &\checkmark &\checkmark & &59.5 &32.3 \\
    GTM-Transformer~\cite{skenderi2024well} &\checkmark &\checkmark &\checkmark &\checkmark &\underline{55,2} &\underline{30,2} \\
    \midrule
    \ours{} (ours) &\checkmark &\checkmark & & &\textbf{54.7} &\textbf{30.1} \\
    \bottomrule
    \end{tabular}
    \label{tab:tab_table_1}
\end{table*} 

This section discusses the quantitative results obtained with \ours{}.
As we can see from Table~\ref{tab:tab_table_1}, \ours{} outperforms all the other state-of-the-art methods without using the information from Google Trends and the textual description of the various samples.
This clearly works in our favor.
Specifically, using Google Trends could worsen the model's performance in our Diffusion-based architecture.
We hypothesize that the information from Google searches may be noisy in some cases and that, with certain samples, this worsens performance.
On the other hand, regarding the description of the model's features, the Diffusion Model probably extracts the information directly from the image features more effectively than obtaining it from the description.
For more information about that, see Section~\ref{subsec:subsec_ablation}.

In particular, we report the comparisons among eight other existing models.
\cite{skenderi2024well} is the most similar in terms of performance to \ours{}.
The most noticeable improvement is in WAPE, which sharply dropped from 55.2 to 54.7, with a slight improvement in MAE as well.
These enhancements in performance offer significant benefits that might not be immediately apparent.
Primarily, as already introduced in Section~\ref{sec:sec_intro}, a Diffusion-based model is always preferable for this task since it should better maintain performance with out-of-distribution objects, ensuring greater stability in practice when used in real-world usage contexts.
Secondly, given our specific architecture, we only need less information to achieve better results.

\subsection{Why the Diffusion Outputs Refinement?} \label{subsec:subsec_qualitative}
\begin{figure*}[t!]
    \centering
    \includegraphics[width=\linewidth]{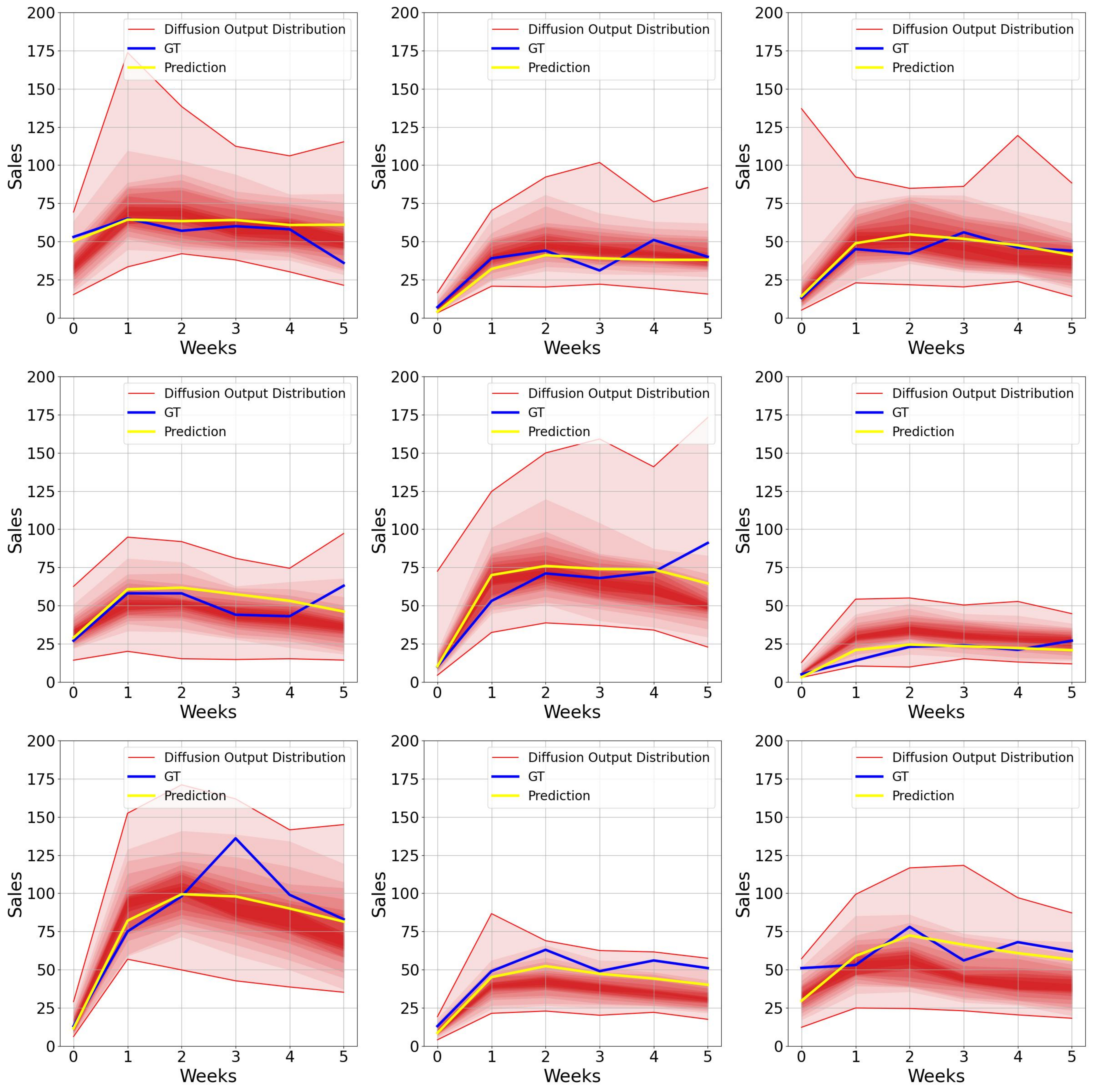}
    \caption{In the figures above are presented some visual representations of the multimodal score-based diffusion model output.
    In particular, the red region represents the output distribution of the Diffusion model given a certain sample.
    The red area is obtained by computing the weekly quantiles among the 50 outputs.
    The Prediction line, on the other hand, is the output of the refinement MLP, \emph{i.e.}, the final prediction.
    The forecasting period is for 6 weeks from the date of release.
    The y-axis shows the number of units sold of a specific garment in the chain's various shops.}
    \label{fig:figure_3}
\end{figure*}

In this section, we motivate the role of the MLP in refining the multimodal score-based diffusion model output forecasting sales.
Starting with the visualization of a few samples in Figure~\ref{fig:figure_3}, it can be seen that the model itself gives as output a distribution of predictions that follows the ground truth very closely.
Therefore, the role of the refinement MLP may seem obsolete, but there are several factors to consider, which instead make it crucial.

Firstly, the model output consists of $N$ predictions; it is then necessary to use some technique to obtain a single final prediction.
Examples might be simply taking the predictions' mean or median.
However, this would result in performance degradation as the ground truth often differs from the distribution's mean or median. 

As a result, we implemented a lightweight MLP network trained to refine the Diffusion Model output, as described in Section~\ref{sec:sec_experiments}.
As we can see from Figure~\ref{fig:figure_3}, specifically in the second and final image, the ground truth does not reside in the densest region of the distribution, and the refinement network very effectively follows its movement away from the median of the diffusion output.

In order to understand how much the refinement model actually helps to improve the performance (and not just predict the mean/median) of the diffusion distribution output, we conducted additional ablative studies reported in Section~\ref{subsec:subsec_ablation} that show that the use of the refinement module is a winning strategy, improving performance considerably without excessively increasing the model complexity. 
We didn't explore how performance might improve with more complex models since our simple MLP already yielded good results.
We plan to delve deeper into this matter and conduct a more thorough analysis in the future.

\subsection{Ablation Studies} \label{subsec:subsec_ablation}
\begin{table*}[!t]
    \centering
    \caption{Table representing the different tests made with the same multimodal score-based diffusion model.
    We tested our model first without the temporal condition and then without images.}
    \begin{tabular}{l|ccc}
    \toprule
    \textbf{Model} & \textbf{WAPE $\downarrow$} & \textbf{MAE $\downarrow$} \\
    \midrule
    \ours{} (ours) without the temporal condition & 56.4 & 31.1 \\
    \ours{} (ours) without the images       & 56.8 & 31.6 \\
    \ours{} (ours) &\textbf{54.7} &\textbf{30.1} \\
    \bottomrule
    \end{tabular}
    \label{tab:tab_table_2}
\end{table*}

We conducted ablative studies to test how well the model performs using different conditioning setups.
This helps us understand which configuration best achieves optimal performance with the Diffusion Model.
It should be noted that the error values of each test done were obtained by running the entire \ours{} pipeline and not just the Diffusion Model.
The results are reported in Table~\ref{tab:tab_table_2}.

Looking at the results, it is clear that conditioning the model with just the images is insufficient.
This is because the features of the dress without information on the season and period in which it is sold is insufficient to predict an accurate sales value.
Indeed, it is not difficult to think that, since the fashion market is a sector strongly influenced by trends, a certain garment may be very fashionable in one season but remain completely unsold in the next.

On the other hand, it is quite straightforward to understand why just the temporal information without any further detail on the item's color, fabric, or shape is insufficient to determine an accurate prediction of the sales. 

It is important to note that for other models such as~\cite{skenderi2024well}, the importance of the various multimodal data types may differ.
In \ours{}, unlike~\cite{skenderi2024well}, conditioning is not processed directly to obtain a prediction but is only used to guide the process of reverse diffusion.
The impact that one type of conditioning can have on different architectures is, therefore, very different from another.

\section{Discussion}
While research on NFPPF has made significant advancements, widespread adoption of the technology still remains far off since predicting signals without historical data is a challenging task requiring integrating information from diverse sources to gauge fashion trends accurately.
We think that this information can be extracted from all those social platforms where people commonly express interests and tastes, such as Instagram and TikTok, allowing breakthrough advancements in NFPPF.

The challenges related to this area are still open, and continuous methodological updates are needed to bridge the gap between theoretical feasibility and practical application.
For instance, the potential impact of garment fit concepts and cross-seasonal comparisons on NFPPF adoption remains largely unexplored, offering promising avenues for innovation.
Hence, Virtual Try-On (VTON) and garment pairing solutions offer promising features for predicting and understanding emerging fashion trends.

In conclusion, while NFPPF technology shows great promise, it is clear that more research and innovation are needed to bring it to practical use.
Integrating data from social media platforms could lead to significant advancements, but challenges like the complexity of garment fit and cross-seasonal comparisons must be addressed.
Continued exploration and development in these areas are essential for moving NFPPF from theory to widespread adoption in the fashion industry, leading to more responsible mass production and reducing the environmental impact of this sector.

\section{Conclusions} \label{sec:sec_conclusions}

In this paper, we propose \ours{}: a novel two-step multimodal Diffusion-based pipeline for NFPPF.
In particular, we first build and train a multimodal score-based diffusion model to provide initial predictions, handling cases with features beyond the training distribution.
Despite the effectiveness of diffusion models, they can sometimes produce inconsistent predictions for the same object.
To tackle this, we generate multiple predictions for each sample using the Diffusion Model and then use them as inputs for the MLP model to make the final prediction.
We tested \ours{} on VISUELLE, the most widely used benchmark for NFPPF, achieving state-of-the-art results.
As a result, this paper further encourages the research of diffusion models for NFPPF.

\paragraph*{\textbf{Future works.}}
We aim to extend our research by exploring the integration of additional data sources, such as customer feedback, to further enhance the predictive accuracy of \ours{}.
Furthermore, we intend to conduct real-world experiments in collaboration with industry partners to validate the effectiveness of our approach in practical settings.
Finally, a more long-term research line is to move from a two-stage pipeline to an end-to-end system.

\section*{Acknowledgement}
This study was carried out within the PNRR research activities of the consortium iNEST (Interconnected North-Est Innovation Ecosystem) funded by the European Union Next-GenerationEU (Piano Nazionale di Ripresa e Resilienza (PNRR) – Missione 4 Componente 2, Investimento 1.5 – D.D. 1058  23/06/2022, ECS\_00000043).
This manuscript reflects only the Authors’ views and opinions. Neither the European Union nor the European Commission can be considered responsible for them.

\bibliographystyle{splncs04}
\bibliography{01_bibi}

\end{document}